\newcommand{\CLASSINPUTtoptextmargin}{0.75in}
\newcommand{\CLASSINPUTbottomtextmargin}{0.905in}

\documentclass[conference,letterpaper,10pt]{IEEEtran}
\IEEEoverridecommandlockouts
\usepackage{graphicx}
\usepackage{booktabs}
\usepackage{cite}
\usepackage{url}

\usepackage{amsmath,amssymb,amsfonts}
\usepackage{mathtools}
\usepackage{bbm}
\usepackage{bbold}
\usepackage{dsfont}
\usepackage{algorithm}
\usepackage{algorithmic}

\usepackage{xcolor}
\usepackage{tikz}
\usetikzlibrary{
arrows,
arrows.meta,
automata,
positioning,
shapes.geometric,
fit,
backgrounds,
calc
}

\usepackage[T1]{fontenc}

\begin{document}

% \title{Multidimensional Suicide-Related Risk Modeling Using MIMIC-IV Electronic Health Records}

\title{Multidimensional Suicide Risk Assessment with Machine Learning using Electronic Health Records}

\author{
\normalsize 
Nathan Scollar\textsuperscript{1},
Kemal Davaslioglu\textsuperscript{1},
Nathan Conger\textsuperscript{1},
Sastry Kompella\textsuperscript{1},
Yalin E. Sagduyu\textsuperscript{1},
and Nathaniel D. Bastian\textsuperscript{2}
\\
\small \textsuperscript{1}Nexcepta, Gaithersburg, MD, USA \\
\small \textsuperscript{2}United States Military Academy, West Point, NY, USA

\thanks{This research was, in part, funded by the U.S. Government. The views and conclusions contained in this document are those of the authors and should not be interpreted as representing the official policies, either expressed or implied, of the U.S. Government.}
}

\maketitle
\thispagestyle{empty}

\begin{abstract}
Suicide prevention remains a critical public health challenge, and electronic health records (EHRs) provide an opportunity to characterize suicide-related risk indicators using documented clinical, behavioral, medical, and social-contextual information. This paper presents a multidimensional machine learning framework to model and estimate suicide-related outcomes using MIMIC-IV EHR data. We construct clinically motivated diagnosis categories and define a composite outcome from documented codes corresponding to suicidal ideation, suicide attempt, prior suicidal behavior, and intentional self-harm. To address severe class imbalance and reduce age-related confounding, we apply age-matched resampling that retains all positive patients and samples comparison-cohort patients from the same age group. Logistic regression, random forest, and XGBoost classifiers are evaluated before and after resampling with isotonic calibration. The calibrated XGBoost model achieves the strongest matched-cohort performance, outperforming the other evaluated classifiers. SHAP-based interpretation shows that diagnostic burden, depression-related features, age, personality-disorder history, social determinants related to violence, and medical comorbidity contribute to elevated predicted probabilities. Our results demonstrate that combining broad EHR-derived features, age-matched cohort construction, calibration, and feature-level interpretation provides a practical framework for retrospective suicide-related risk modeling. %provides a practical framework for suicide-related risk modeling and prediction.
\end{abstract}


\section{Introduction}

Electronic health records (EHRs) provide a valuable foundation for health-risk modeling by capturing clinical, behavioral, and contextual information across patient encounters. Advances in machine learning make it possible to analyze these complex and heterogeneous data to support predictive healthcare analytics and earlier clinical assessment. Suicide prevention is a particularly important application: in 2023, nearly 50,000 Americans died by suicide, while 12.8 million seriously considered suicide, 3.7 million made a plan, and 1.5 million attempted suicide \cite{cdc}. U.S. suicide rates increased by 37\% between 2000 and 2018, briefly declined between 2018 and 2020, and returned to their peak by 2022 \cite{cdc}. These trends highlight the need for data-driven methods that can characterize suicide-related risk indicators and support more informed clinical assessment.

%This paper investigates suicide-related risk modeling and prediction using MIMIC-IV

This paper investigates retrospective suicide-related risk modeling using MIMIC-IV, a large-scale EHR dataset containing emergency department and intensive care unit records from over 200,000 patients \cite{johnson2023mimic}. Building on prior machine learning approaches for suicide risk assessment that use structured EHR variables, temporal information, and ensemble learning \cite{TemporallyInformedRandomForests,SwedishNational}, our objective is to develop a retrospective modeling framework that integrates psychiatric, behavioral, medical, and social-determinant-related features associated with documented suicide-related outcomes. We first perform odds-ratio analysis to identify diagnosis categories with strong univariate associations with the composite outcome, then train and calibrate logistic regression, random forest, and XGBoost classifiers under both imbalanced and age-matched cohort settings. To support interpretability, we use SHAP-based feature attribution to examine which EHR-derived factors contribute most strongly to elevated predicted probabilities \cite{lundberg2017unified}. Although mental health diagnoses provide essential information, this framework is designed to capture broader patterns across the documented health profilse that may not be represented by psychiatric variables alone.

The novel contributions of this paper are as follows:

\begin{enumerate}
\item  \textbf{Multidimensional EHR Feature Framework:} We construct 53 clinically motivated diagnosis categories spanning psychiatric conditions, substance abuse, cognitive and behavioral factors, physical health conditions, and social determinants of health to support retrospective suicide-related risk modeling.
\item  \textbf{Age-Matched Cohort Construction for Class Imbalance:} We introduce an age-matched resampling strategy that retains all positive patients while selecting three negative patients from the same age group for each positive patient, ensuring a consistent 25\% positive rate across age groups and reducing confounding associated with age-dependent outcome prevalence.
\item \textbf{Calibrated Model Evaluation and Interpretation:} We develop an evaluation framework that compares logistic regression, random forest, and XGBoost classifiers before and after age-matched resampling, applies isotonic calibration, and assesses robustness across 20 random seeds. The calibrated XGBoost model achieves the strongest matched-cohort performance, with an F1 score of 0.81 and an Area Under the Precision-Recall Curve (AUPRC) of 0.89. Across the 20-seed robustness analysis, XGBoost maintains stable performance, with an average F1 score of 0.81 and AUPRC of 0.89. We further use SHAP-based analysis to interpret predicted probabilities and identify the EHR-derived features that contribute most strongly to model predictions.
\end{enumerate}

The remainder of this paper is organized as follows. Section~\ref{sec:related_work} reviews related work. Sections~\ref{sec:data_preprocessing}--\ref{sec:resampling} describe data preprocessing, exploratory analysis, and age-matched resampling. Section~\ref{sec:modeling} presents the modeling framework, Section~\ref{sec:experiments} reports the experimental results, and Section~\ref{sec:conclusion} concludes the paper.

\section{Related Work}\label{sec:related_work}

Prior work has applied machine learning methods to suicide-related risk modeling, including a balanced random forest approach that embeds temporal information within each tree to support suicide risk assessment \cite{TemporallyInformedRandomForests}, ensemble models trained on Swedish registry data that combine penalized logistic regression, random forests, gradient boosting, and neural networks \cite{SwedishNational}, and interpretable CART models for suicide-attempt and self-harm risk after medical hospitalization \cite{UCLA}. Clinical narratives have also been analyzed to identify warning signs preceding suicide attempts among adults with substance use problems \cite{Warning}. These studies demonstrate the value of structured EHR variables, temporal information, ensemble learning, interpretable models, and clinical narratives.

A related body of work focuses on veterans and military populations, including discrete-time survival models for long-horizon suicide prediction \cite{ArmySuicidePrediction}, STARRS-LS models based on baseline and post-service surveys \cite{STARRS}, age-stratified logistic regression for preventive intervention \cite{VA}, and REACH VET evaluations for legal-involved veterans \cite{REACHVET}. These efforts highlight the operational importance of suicide-risk modeling in mission-relevant populations and motivate generalizable methods that can be validated on large-scale EHR data.

Recent network-analysis work further supports a multidimensional view of suicide-related risk by jointly examining sociodemographic, clinical, and psychopharmacological factors in psychiatric cohorts with multi-affected families \cite{torino2026suicide}. Although focused on association modeling rather than large-scale EHR prediction, this work reinforces the importance of jointly modeling clinical, behavioral, and contextual factors. Our complementary work in BEACON-SP \cite{beaconsp} investigates ontology-grounded GraphRAG for clinician-facing suicide-prevention decision support, whereas the present study focuses on retrospective EHR-based risk modeling.

This paper differs from prior work by combining broad diagnosis-category engineering, age-matched cohort construction, probability calibration, robustness testing across random resampling seeds, and SHAP-based feature interpretation within a single MIMIC-IV EHR modeling framework. Rather than focusing only on a specific population, a single model family, temporal structure, or narrative text, our approach evaluates multiple calibrated classifiers on multidimensional EHR-derived features and explicitly examines how age-matched resampling affects performance under severe class imbalance.

\section{Data Pre-Processing}\label{sec:data_preprocessing}
\begin{figure}[t!]
    \centering
    \includegraphics[width=\columnwidth]{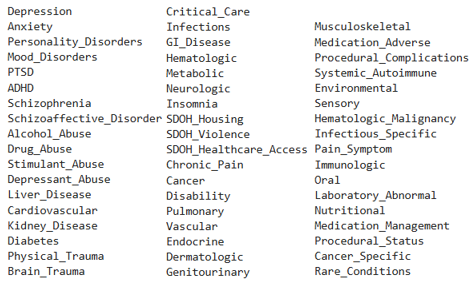}
    \caption{The 53 health-related categories utilized for creating features.}
    \label{fig:MimicCategories}
\end{figure}

The MIMIC-IV dataset contains data from the emergency room and ICU visits of over 200,000 patients, including over 6.3 million recorded diagnoses. Out of these diagnoses, there are over 27,000 unique clinical diagnoses.  The 53 clinical categories are constructed using keyword-based mappings that group diagnoses with related clinical meanings, including diagnoses expressed using slightly different terminology. These categories span psychiatric conditions, substance abuse, cognitive and behavioral factors, social determinants of health (SDoH), and other physical health conditions. The complete category set is shown in Fig~\ref{fig:MimicCategories}. For each clinical category, we generate two predictor features: (i) a binary indicator representing whether the patient has at least one diagnosis in that category and (ii) a count representing the number of recorded diagnoses within the category. We also include one-hot encoded age-group and sex indicators, together with each patient's total number of recorded diagnoses. The age groups are 18--25, 26--35, 36--45, 46--55, 56--65, 66--75, 76--85, and 86+. The resulting feature vector contains 117 predictors: 106 category-derived features, 8 age-group indicators, 2 sex indicators, and total diagnostic count.

The binary outcome is constructed from documented suicide-related diagnosis and external-cause codes. ICD-9-CM codes E950--E959 identify suicide and self-inflicted injury events, while ICD-9-CM code V62.84 identifies suicidal ideation. ICD-10-CM codes X71--X83 identify intentional self-harm by specified mechanisms other than poisoning and asphyxiation; ICD-10-CM code T14.91 identifies suicide attempt; ICD-10-CM code R45.851 identifies suicidal ideation; and ICD-10-CM code Z91.51 identifies a personal history of suicidal behavior. We also include intentional self-harm poisoning codes within ICD-10-CM ranges T36--T50 when the intent character indicates intentional self-harm.

Patients with at least one diagnosis or external-cause code satisfying this composite suicide-related outcome definition are assigned to the positive, or high-risk, cohort. Patients without a documented code satisfying this outcome definition are assigned to the negative, or comparison, cohort. The terms \emph{high-risk cohort} and \emph{control cohort} are used as operational EHR-derived study labels and should not be interpreted as prospective clinical risk determinations. Outcome-defining codes were used only for label construction and were excluded from the predictor feature set.

\section{Exploratory Data Analysis}\label{sec:eda}
Initial exploratory analysis was performed using odds ratios to identify categories associated with the composite suicide-related outcome. For a binary category feature, the odds ratio quantifies the odds of that category being present in the high-risk cohort relative to the control cohort. Odds ratios greater than 1 indicate that the category is more prevalent in the high-risk cohort, whereas odds ratios less than 1 indicate that the category is more prevalent in the control cohort.

Table~\ref{tab:Top10OddsRatios} reports the ten categories with the largest odds ratios. Among the SDoH-related categories, SDOH Housing primarily represents diagnoses related to homelessness or housing instability, whereas SDOH Violence captures diagnoses related to abuse or violence exposure, including physical or sexual abuse. The High-Risk Cohort Rate column reports the percentage of patients satisfying the composite suicide-related outcome definition who have at least one diagnosis in the corresponding category. The Control Cohort Rate column reports the corresponding percentage among patients without documented positive outcome codes.

\begin{table}[t!]
\caption{Categories with the ten largest odds ratios for the composite suicide-related outcome.}
\label{tab:Top10OddsRatios}
\centering
\resizebox{\columnwidth}{!}{%
\begin{tabular}{cccc}
\toprule
Feature & High-Risk Cohort Rate & Control Cohort Rate & Odds Ratio \\
\midrule
SDOH Violence & 29.6\% & 1.2\% & 34.62 \\
Mood Disorders & 18.8\% & 0.9\% & 26.13 \\
PTSD & 21.3\% & 1.6\% & 17.03 \\
SDOH Housing & 17.6\% & 1.3\% & 16.62 \\
Depression & 75.6\% & 16.6\% & 15.53 \\
Personality Disorders & 26.7\% & 2.8\% & 12.60 \\
Schizoaffective Disorder & 5.9\% & 0.6\% & 10.89 \\
Depressant Abuse & 13.9\% & 1.7\% & 9.49 \\
ADHD & 8.7\% & 1.3\% & 7.15 \\
Schizophrenia & 5.8\% & 0.9\% & 6.66 \\
\bottomrule
\end{tabular}
}

\end{table}

Table~\ref{tab:Bottom10OddsRatios} shows the 10 features with the lowest odds ratios. It is worth noting that the difference between Cancer and Cancer Specific is that Cancer encompasses more common and general cancer-related diagnoses, whereas Cancer Specific contains rarer and more specific cancer-related diagnoses.

\begin{table}[t!]
\caption{Categories with the ten smallest odds ratios for the composite suicide-related outcome.}
\label{tab:Bottom10OddsRatios}
\centering
\resizebox{\columnwidth}{!}{%
\begin{tabular}{cccc}
\toprule
Feature & High-Risk Cohort Rate & Control Cohort Rate & Odds Ratio \\
\midrule
Cancer & 7.9\% & 22.2\% & 0.30 \\
Vascular & 13.1\% & 31.3\% & 0.33 \\
Cardiovascular & 35.9\% & 61.9\% & 0.34 \\
Hematologic Malignancy & 0.8\% & 2.3\% & 0.35 \\
Endocrine & 29.1\% & 52.8\% & 0.37 \\
Procedural Status & 7.7\% & 16.6\% & 0.42 \\
Kidney Disease & 15.0\% & 28.5\% & 0.44 \\
Cancer Specific & 1.7\% & 3.5\% & 0.47 \\
Hematologic & 24.6\% & 40.1\% & 0.49 \\
Diabetes & 13.0\% & 22.9\% & 0.50 \\
\bottomrule
\end{tabular}
}

\end{table}

\section{Age-Matched Resampling}\label{sec:resampling}

The original dataset is highly imbalanced, containing 11,197 patients in the high-risk cohort and 212,255 patients in the control cohort, corresponding to a positive outcome rate of 5.0\%. To construct an age-balanced analytical cohort, we retain all patients in the high-risk cohort and sample three comparison-cohort patients from the same age group for each high-risk-cohort patient. This procedure produces a 3:1 comparison-to-high-risk ratio and a 25\% positive outcome rate within each age group and across the matched cohort overall. Because this resampled cohort does not preserve the original outcome prevalence, results obtained on this cohort characterize model performance under age-matched evaluation conditions rather than expected clinical deployment performance.

We also conducted preliminary experiments with Synthetic Minority Over-Sampling Technique (SMOTE) \cite{chawla2002smote}, Adaptive Synthetic Sampling Approach for Imbalanced Learning (ADASYN) \cite{he2008adasyn}, and decision-threshold tuning. These approaches were not retained for the primary analysis because age-matched cohort construction provided the strongest validation performance while explicitly controlling the age distribution of the control cohort. In addition, class weighting was applied during model training: logistic regression and random forest used balanced class weight, and XGBoost used the balance of positive and negative weights based on the training-cohort class ratio.

\section{Suicide-Related Risk Modeling}\label{sec:modeling}

We evaluate three supervised classifiers: logistic regression, random forest, and XGBoost. Each model is hyperparameter tuned and calibrated using an age-group-stratified 70/30 train/test split, with hyperparameters selected by average F1 score from five-fold cross-validation on the training set. 

We organize the analysis into three experiments. Experiment 1 compares the three classifiers before and after age-matched resampling using a single seed, and further characterizes the best-performing model through feature importance and age-group performance. Experiment 2 quantifies the stability of the best-performing model across 20 random seeds. 
%Experiment 3 conducts a realistic-prevalence analysis on comparison-cohort patients held out from matched training, 
The third experiment complements the matched-cohort evaluation with a held-out comparison-cohort analysis, examining which documented features drive elevated predicted probabilities. The matched-cohort experiments quantify performance under controlled age-group representation, whereas the held-out analysis examines model behavior among comparison-cohort patients excluded from matched training.

\begin{table}[t!]
\centering
\caption{Hyperparameters tuned and the values tested for each model.}
\label{tab:HyperparameterTuning}
\resizebox{\columnwidth}{!}{%
\begin{tabular}{ccc}
\toprule
Model & Feature & Values Tested  \\
\midrule
Logistic Regression & C & 0.01, 0.1, 1, 10 \\
Logistic Regression & Penalty & L1, L2 \\
Random Forest & num estimators & 200, 400 \\
Random Forest & max depth & None, 20, 30 \\
Random Forest & min samples split & 2, 5 \\
XGBoost & num estimators & 100, 200, 300 \\
XGBoost & max depth & 5, 7, 9 \\
XGBoost & learning rate & 0.01, 0.05, 0.1, 0.2 \\
\bottomrule
\end{tabular}
}

\end{table}

\subsection{Hyperparameter Tuning}
Each model is hyperparameter tuned using average F1 score across five-fold cross-validation on the training set as the selection metric. The full set of tuned hyperparameters and the corresponding values tested are presented in Table~\ref{tab:HyperparameterTuning}.

\subsection{Model Calibration}
Each classifier is calibrated using CalibratedClassifierCV from the scikit-learn library. Calibration aligns predicted probabilities with observed outcome frequencies in the evaluation cohort. We use isotonic calibration as it does not impose a fixed sigmoid mapping between classifier scores and calibrated probabilities. 

\begin{figure}[t!]
    \centering
    \includegraphics[width=\columnwidth,height=2.5in]{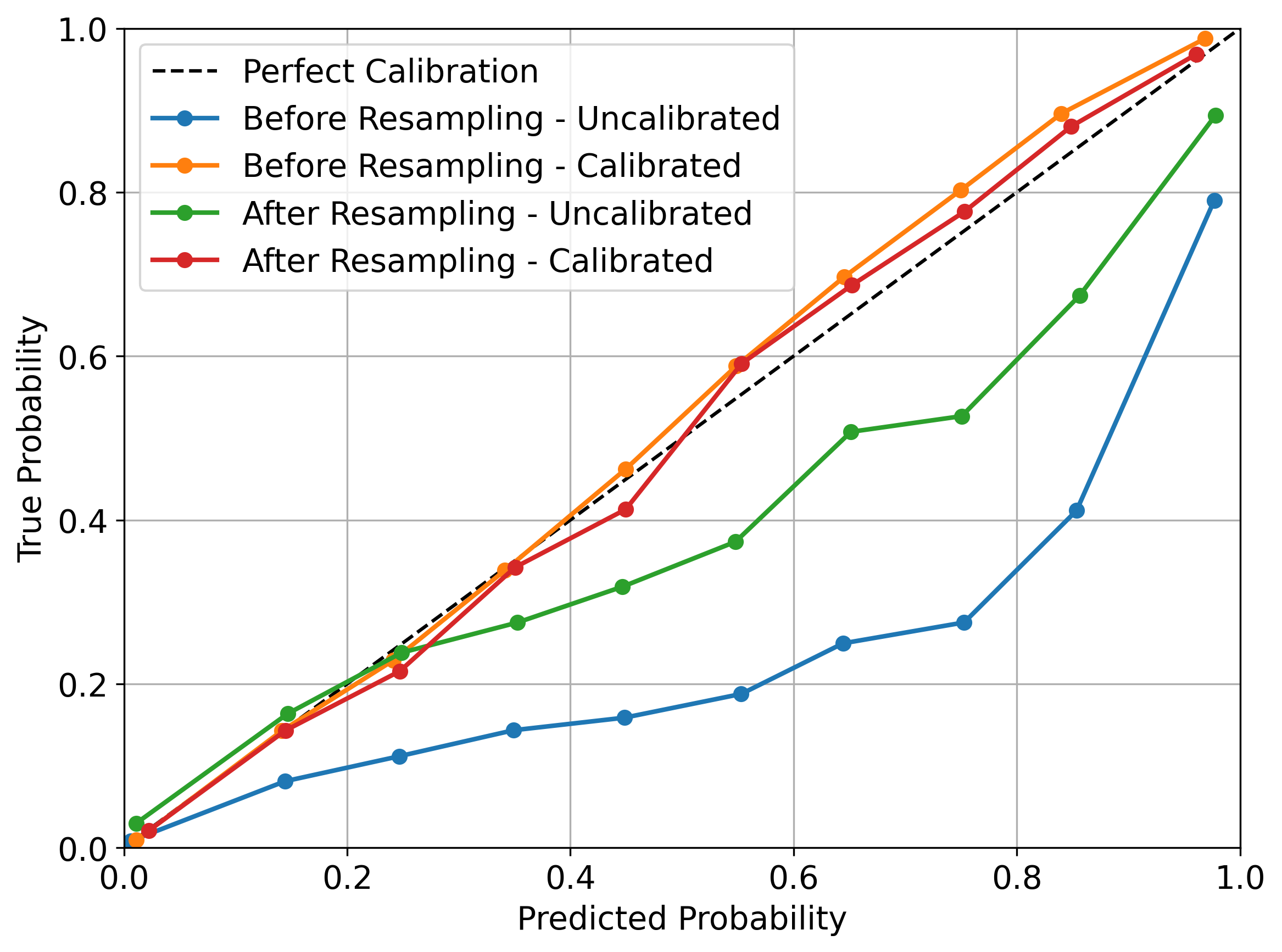}
    \caption{The impact of isotonic model calibration.}
    \label{fig:CalibrationComparison}
\end{figure}

Fig.~\ref{fig:CalibrationComparison} compares the calibrated and uncalibrated XGBoost models. Prior to calibration, the XGBoost calibration curves fall below the ideal calibration line, indicating that predicted probabilities exceed observed positive outcome rates over much of the probability range. Isotonic calibration substantially improves agreement between predicted probabilities and observed rates within both evaluated cohorts. Importantly, calibration on the age-matched cohort reflects its constructed 25\% positive outcome prevalence and should not be interpreted as calibration under real-world deployment prevalence.

\section{Experiments and Results}\label{sec:experiments}

\subsection{Experiment 1: Classifier Comparison Results}
The first experiment compares logistic regression, random forest, and XGBoost using a single age-group-stratified 70/30 split. Each model is first evaluated on the original imbalanced cohort and then re-evaluated after age-matched resampling, isolating the effect of cohort construction on classification performance.

The full classifier results are shown in Table~\ref{tab:ClassifierResults}. The XGBoost model had the strongest results both before and after age-matched resampling, with a 0.6785 F1 score and a 0.7573 Area Under the Precision-Recall Curve (AUPRC) prior to age-matched resampling, compared to a 0.8108 F1 score and a 0.8891 AUPRC after age-matched resampling. All three models achieve higher F1 and AUPRC under the age-matched 25\% prevalence setting. %These values should not be interpreted as direct performance improvements over the original-prevalence cohort as the evaluation distributions different.

%The biggest takeaway here was the significant improvements across all three models due to age-matched resampling, as the logistic regression and random forest models also saw major improvements to their F1 scores and AUPRC after age-matched resampling.

\begin{table}[t!]
\caption{Classification results for the composite suicide-related outcome before and after age-matched resampling.}
\label{tab:ClassifierResults}
\centering
\resizebox{\columnwidth}{!}{%
\begin{tabular}{cccccc}
\toprule
Model & Precision & Recall & F1 & AUPRC & AUROC \\
\midrule
Logistic Regression (Before) & 0.7282 & 0.4475 & 0.5543 & 0.6567 & 0.9543 \\
Logistic Regression (After) & 0.7781 & 0.7782 & 0.7816 & 0.8377 & 0.9343  \\
Random Forest (Before) & 0.7429 & 0.5350 & 0.6220 & 0.6896 & 0.9614 \\
Random Forest (After) & 0.8157 & \textbf{0.7812} & 0.7981 & 0.8701 & 0.9501 \\
XGBoost (Before) & 0.8175 & 0.5799 & 0.6785 & 0.7573 & 0.9603 \\
XGBoost (After) &  \textbf{0.8434} & 0.7806 & \textbf{0.8108} & \textbf{0.8891} & \textbf{0.9533} \\
\bottomrule
\end{tabular}
}
\end{table}

\begin{figure}[t!]
    \centering
    \includegraphics[width=0.95\columnwidth]{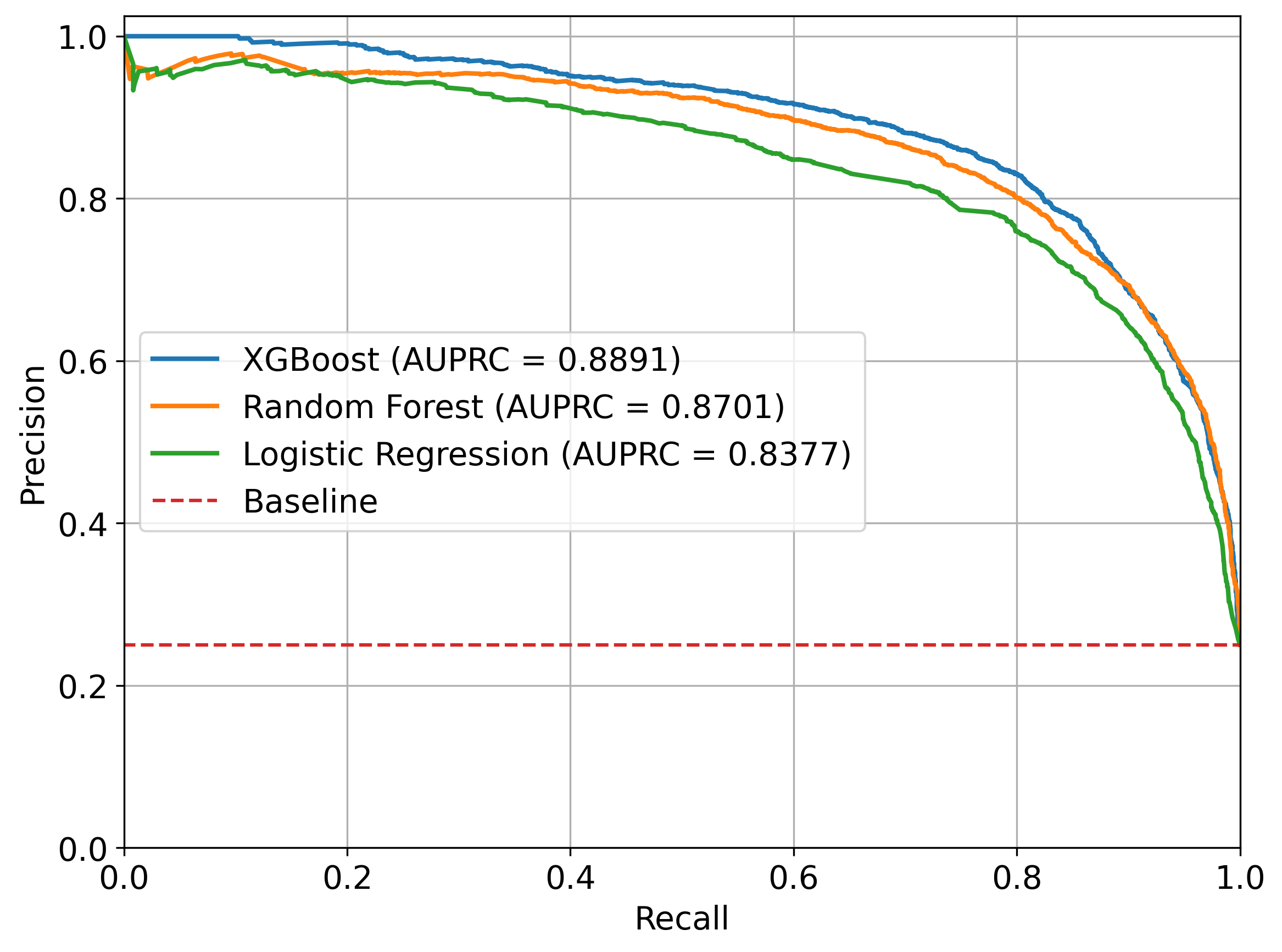}
    \caption{Precision-recall curves of all three models after age-matched resampling.}
    \label{fig:PrecisionRecallCurve}
\end{figure}

Fig.~\ref{fig:PrecisionRecallCurve} presents the precision-recall curves for the three classifiers following age-matched resampling. XGBoost achieves the strongest overall precision-recall performance, with an average precision of 0.889, outperforming logistic regression and random forest across much of the recall range. This result indicates that XGBoost provides a better trade-off between precision and recall for identifying patients in the positive outcome cohort under the age-matched evaluation setting.

\subsubsection{XGBoost Feature Importance}
\begin{figure}[t!]
    \centering
    \includegraphics[width=\columnwidth]{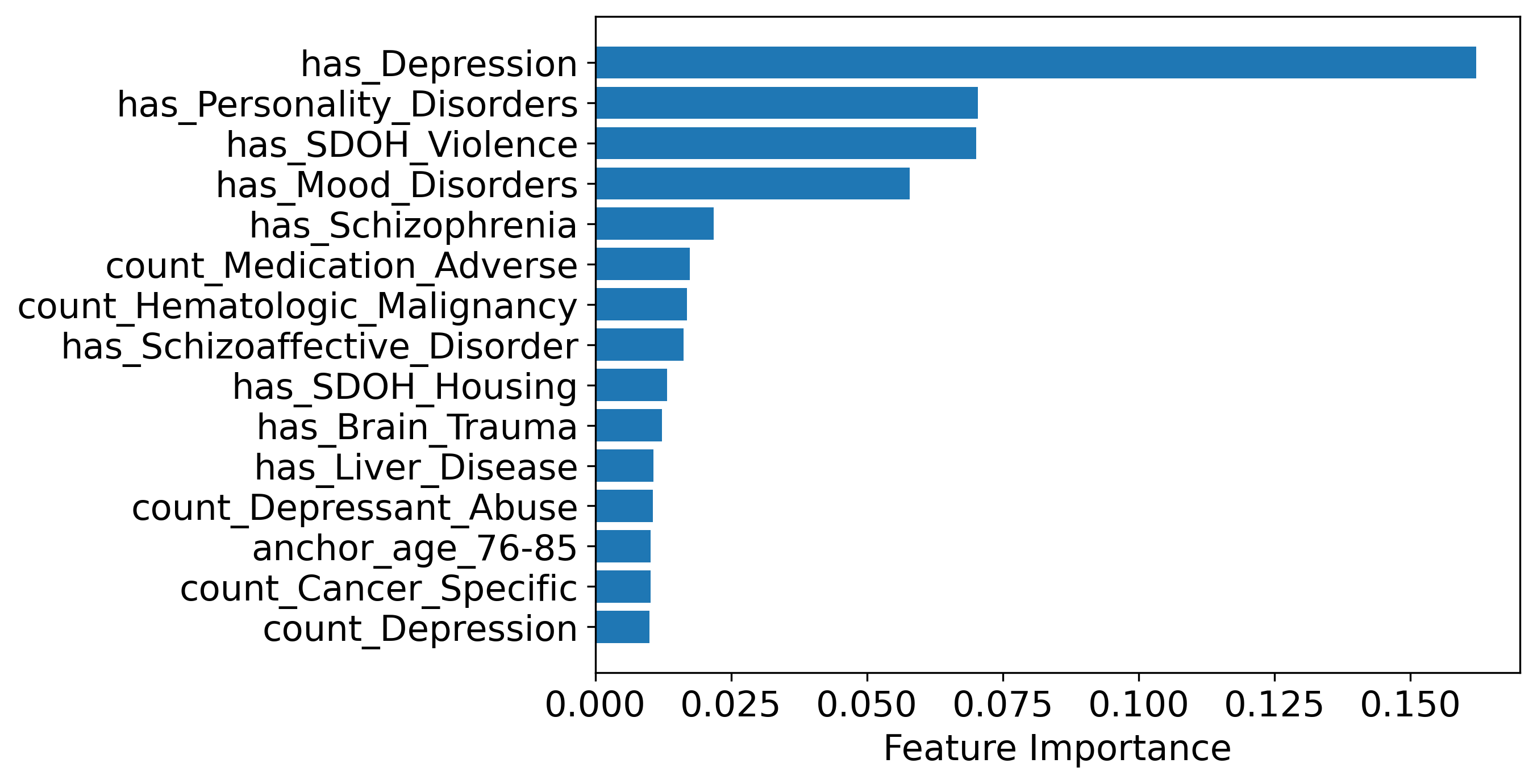}
    \caption{Top 15 XGBoost feature importance values for classification of the composite suicide-related outcome after age-matched resampling.}
    \label{fig:FeatureImportance}
\end{figure}

Model-derived feature importance values from XGBoost were examined to identify the factors contributing most strongly to classification decisions in the age-matched cohort. These values describe relative contributions within the trained classifier and should not be interpreted as causal effects or independent clinical risk factors. The most influential feature was the presence of depression, followed by personality disorders, social determinants of health related to violence, mood disorders, and schizophrenia. The top 15 features for the calibrated XGBoost model after age-matched resampling are shown in Fig.~\ref{fig:FeatureImportance}.

\subsubsection{Performance by Age Group}
Table~\ref{tab:AgeGroupResults} reports the test size, precision, recall, and F1 score for the calibrated XGBoost classifier within each age group. Because age-matched resampling enforces a 25\% positive outcome rate within every age group, differences in these metrics reflect differences in classification difficulty across age groups under controlled class representation rather than differences in outcome prevalence within the evaluated cohort.

Performance is strongest for patients between 18 and 55 years of age, with F1 scores exceeding 0.81 across the four youngest age groups. Performance decreases for older age groups, with the largest declines occurring among patients aged 76 years or older. This decrease is driven primarily by lower recall where the model identifies a smaller fraction of positive outcome cases in older age groups even under matched evaluation conditions. These findings should be interpreted cautiously because the oldest age groups have substantially smaller evaluation sample sizes. Additional evaluation using larger older-adult cohorts and untouched outcome prevalence is needed before drawing conclusions regarding age-specific clinical utility.

\begin{table}[t!]
\centering
\caption{Age-group-specific results for the calibrated XGBoost classifier within the age-matched evaluation cohort.}
\label{tab:AgeGroupResults}
\resizebox{\columnwidth}{!}{%
\begin{tabular}{ccccc}
\toprule
Age Group & Test Size & Precision & Recall & F1 Score \\
\midrule
18--25 & 4601 & 0.8418 & 0.8330 & 0.8374 \\
26--35 & 2696 & 0.8547 & 0.8027 & 0.8279 \\
36--45 & 2167 & 0.8466 & 0.7841 & 0.8142 \\
46--55 & 1993 & 0.8620 & 0.7651 & 0.8106 \\
56--65 & 1207 & 0.8171 & 0.6953 & 0.7513 \\
66--75 & 463  & 0.7917 & 0.6552 & 0.7170 \\
76--85 & 241  & 0.7272 & 0.4000 & 0.5161 \\
86+    & 69   & 0.8750 & 0.4118 & 0.5600 \\
\bottomrule
\end{tabular}
}
\end{table}

\subsection{Experiment 2: Robustness Results}
To evaluate sensitivity to the random selection of comparison-cohort patients, we repeat age-matched resampling and evaluation of the calibrated XGBoost classifier with optimal hyperparameters across 20 randomly selected seeds. The seeds are drawn without replacement from the integers between 1 and 100. For each seed, the full age-matched resampling process is performed, with the seed determining both which comparison-cohort patients are retained and the train/test split. For every seed we record accuracy, precision, recall, F1 score, AUPRC, Area Under the Receiver Operating Characteristic (AUROC), and we report the average and 95\% confidence interval of each metric.

Table~\ref{tab:RobustnessExperiment} reports the average performance and corresponding 95\% confidence intervals. The calibrated XGBoost classifier achieved an average F1 score of 0.8062 with a 95\% confidence interval of [0.8042, 0.8083], and an average AUPRC of 0.8904 with a 95\% confidence interval of [0.8888, 0.8921]. The narrow confidence intervals across evaluation metrics indicate that classifier performance within the age-matched cohort is stable across the evaluated random resampling seeds.

\begin{table}[t!]
\centering
\caption{Performance of the calibrated XGBoost classifier with age-matched resampling across 20 random seeds.}
\label{tab:RobustnessExperiment}
\resizebox{\columnwidth}{!}{%
\begin{tabular}{cccc}
\toprule
Metric & Average Value & 95\% CI Lower Bound & 95\% CI Upper Bound \\
\midrule
Precision & 0.8368 & 0.8335 & 0.8401 \\
Recall & 0.7779 & 0.7744 & 0.7814 \\
F1 Score & 0.8062 & 0.8042 & 0.8083 \\
AUPRC & 0.8904 & 0.8888 & 0.8921 \\
AUROC & 0.9545 & 0.9539 & 0.9552 \\
\bottomrule
\end{tabular}
}
\end{table}

\begin{table}[t!]
\caption{Top 10 features by mean absolute SHAP value among high-scoring patients in the held-out control cohort.}
\label{tab:ComparisonCohortSHAP}
\centering
\resizebox{\columnwidth}{!}{
\begin{tabular}{cccc}
\toprule
Feature & Mean Abs. SHAP & SD & CV \\
\midrule
Total diagnoses & 0.2176 & 0.0115 & 0.0528 \\
Has depression & 0.0900 & 0.0077 & 0.0858 \\
Count depression & 0.0818 & 0.0055 & 0.0674 \\
Age 76--85 & 0.0593 & 0.0128 & 0.2164 \\
Has Personality Disorders & 0.0570 & 0.0070 & 0.1220 \\
Has SDOH Violence & 0.0496 & 0.0093 & 0.1869 \\
Age 66--75 & 0.0437 & 0.0077 & 0.1754 \\
Count Cardiovascular & 0.0435 & 0.0054 & 0.1237 \\
Count GI Disease & 0.0384 & 0.0037 & 0.0968 \\
Has Mood Disorders & 0.0335 & 0.0067 & 0.1991 \\
\bottomrule
\end{tabular}
}
\end{table}

\subsection{Experiment 3: Held-Out Comparison-Cohort Results}
The third experiment complements matched-cohort evaluation with a held-out comparison cohort excluded from training. For each seed, the training cohort contains all high-risk patients and three age-matched comparison patients per positive case; the remaining comparison patients, none with documented positive outcome codes, form the held-out set. Calibrated XGBoost is trained on the matched cohort and applied to the held-out cohort. We then examine the top 5\% of predicted probabilities to identify features associated with elevated scores among patients without documented positive outcomes.
%The third experiment complements the matched-cohort evaluation with a held-out comparison-cohort analysis using comparison-cohort patients excluded from matched training. For each seed, the training cohort contains all high-risk-cohort patients together with three randomly sampled comparison-cohort patients from the same age group for each positive patient. The remaining comparison-cohort patients, none of which contain documented positive outcome codes, are retained as a held-out evaluation set. The calibrated XGBoost classifier is trained on the matched training cohort, and predicted probabilities are computed for the held-out control cohort. We then examine the top 5\% of predicted probabilities to characterize the features contributing to elevated model scores among patients without documented positive outcome codes.

The primary interpretability metric is the mean absolute SHAP value averaged across the 20 experimental seeds, which summarizes the magnitude of each feature's contribution to the model output independent of direction. We also report the standard deviation across seeds and the coefficient of variation (CV), where smaller CV values indicate more consistent feature contribution estimates.

\begin{figure}[t!]
\centering
\includegraphics[width=\columnwidth]{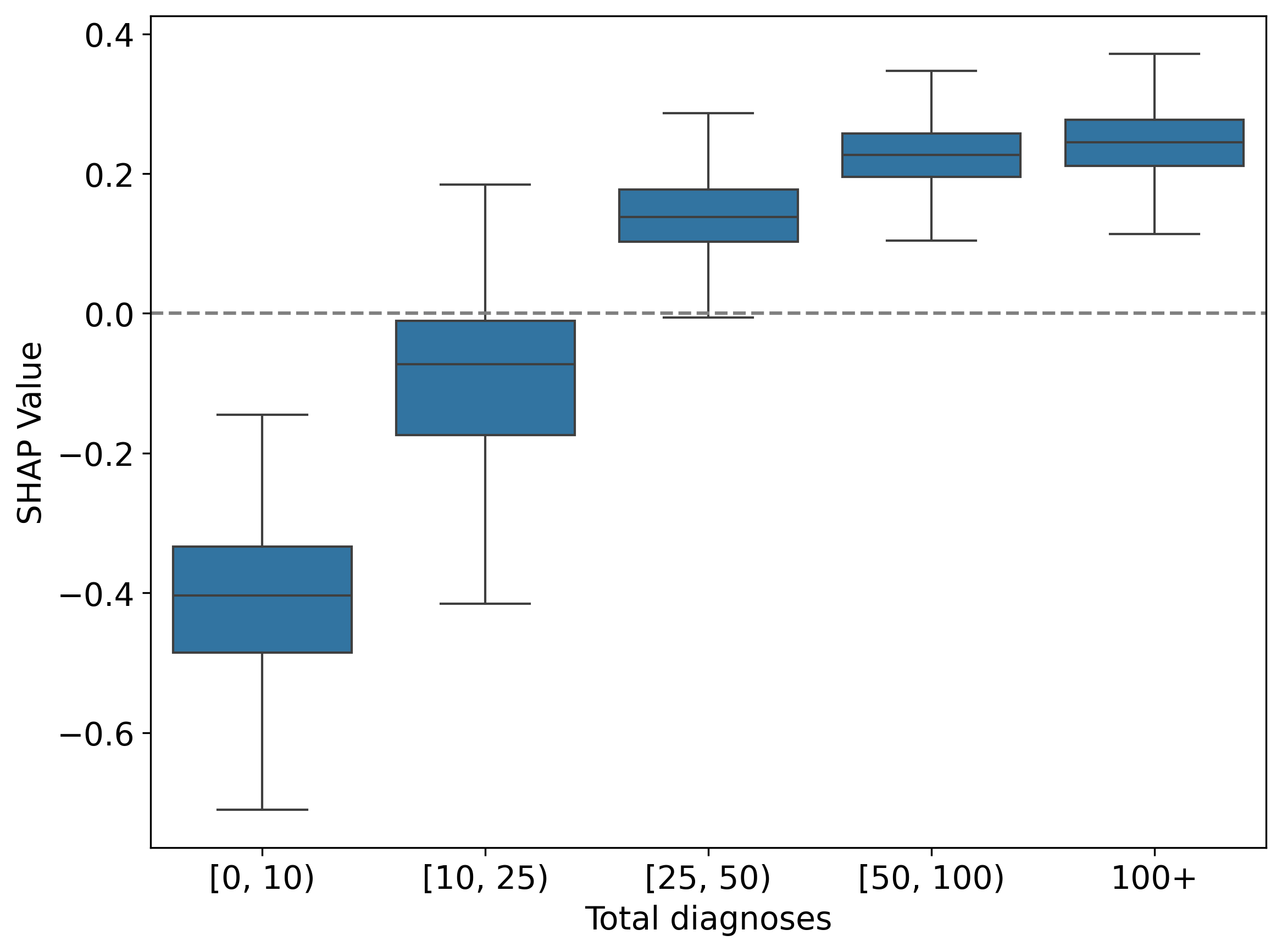}
\caption{Distribution of SHAP values for total diagnoses among high-scoring patients in the held-out control cohort.}
\label{fig:shapBoxplot}
\end{figure}

Table~\ref{tab:ComparisonCohortSHAP} shows the ten features with the largest mean absolute SHAP values among high-scoring patients in the held-out comparison cohort. The feature \emph{total diagnoses} is the dominant contributor to elevated predicted probabilities, with a mean absolute SHAP value of 0.2176 and a low CV of 0.0528. Several SHAP-identified contributors, including depression-related features, personality-disorder history, SDoH-related violence, and mood disorders, are consistent with the strongest associations observed in the odds-ratio analysis. This agreement suggests that key psychiatric and social-contextual categories remain important across both univariate association analysis and multivariate model interpretation. At the same time, SHAP also highlights diagnostic burden, older age indicators, and medical comorbidities such as cardiovascular and gastrointestinal disease, indicating that the model integrates broader clinical context beyond the categories with the largest individual odds ratios. Fig.~\ref{fig:shapBoxplot} shows the distribution of SHAP values for total diagnoses among these high-scoring patients. Note that these results should not be interpreted as demonstrating confirmed false positives. A patient assigned to the control cohort may lack a documented positive outcome code in the available EHR record while still exhibiting unobserved, undocumented, or future suicide-related risk. Instead, this analysis characterizes which documented features cause the model to assign elevated scores among patients without recorded positive outcomes. % codes.

\section{Conclusion}\label{sec:conclusion}

This paper presented a multidimensional machine learning framework for retrospective suicide-related risk modeling using MIMIC-IV EHR data. We constructed clinically motivated diagnosis categories spanning psychiatric, medical, behavioral, and social-determinant-related factors and defined a composite outcome from documented codes for suicidal ideation, suicide attempt, prior suicidal behavior, and intentional self-harm. To address class imbalance and reduce age-related confounding, we applied age-matched cohort construction and evaluated logistic regression, random forest, and XGBoost with isotonic calibration. All classifiers achieved higher F1 and AUPRC under age-matched evaluation conditions, with calibrated XGBoost achieving the strongest matched-cohort results. Robustness testing across 20 random seeds showed stable performance, and SHAP-based interpretation identified diagnostic burden, depression-related features, age, personality-disorder history, SDoH-related violence, and medical comorbidity as major contributors to elevated predicted probabilities. The evaluation results show that broad EHR-derived features, age-matched cohort construction, calibration, and feature-level interpretation provide a practical framework for suicide-related risk modeling. Future work will extend this approach toward patient-level encounter-sequence and time-to-event modeling to capture how suicide-related risk evolves over a patient's documented history.

\bibliographystyle{ieeetr}
\bibliography{references}

\end{document}